\documentclass[conference,letterpaper]{IEEEtran}
\IEEEoverridecommandlockouts

\usepackage{amssymb}
\usepackage{amsmath}
\usepackage{mathtools}
\usepackage{bbm}
\usepackage{bm}
\usepackage{graphicx}
\usepackage{xcolor}
\usepackage{url}
\usepackage{nicefrac}
\usepackage{url}
\usepackage{bm}
\usepackage[normalem]{ulem}
\usepackage{tabularx}
\usepackage{subcaption}
\usepackage{adjustbox}
\usepackage{hyperref}
\usepackage[linesnumbered]{algorithm2e}

\usepackage{array}
\usepackage{booktabs}

\usepackage{harmony}

\usepackage{tikz}
\usepackage{pgfplots}
\pgfplotsset{compat=1.3}
\usepackage{filecontents}
\usetikzlibrary{matrix}
\usetikzlibrary{calc}
\usetikzlibrary{arrows}
\usetikzlibrary{patterns}
\usetikzlibrary{plotmarks}
\usetikzlibrary{intersections} 
\usetikzlibrary{decorations.pathmorphing}
\usetikzlibrary{decorations.pathreplacing}
\usetikzlibrary{decorations.markings,arrows} 
\usetikzlibrary{decorations.shapes}
\usetikzlibrary{fit}
\usetikzlibrary{backgrounds}
\usetikzlibrary{trees}

\newcommand{\norm}[1]{\left\lVert#1\right\rVert}

\newcommand{\encoder}[1]{f_{\text{enc}}\!\left(#1\right)}
\newcommand{\decoder}[1]{g_{\text{dec}}\!\left(#1\right)}

\newcommand{\expectationwrt}[2]{\mathbb{E}_{#1}\!\left[#2\right]}
\newcommand{\kl}[2]{D_\mathrm{KL}\!\left[#1 \;\middle\|\; #2 \right]}
\newcommand{\klelem}[2]{\mathrm{KL}_{\mathrm{elem}}\!\left[#1 \;\middle\|\; #2 \right]}
\newcommand{\constraint}[1]{\mathcal{C}\!\left(#1\right)}

\usepackage[colorinlistoftodos]{todonotes}
\usepackage{soul}

\definecolor{ugent-blauw}{RGB}{30,100,200}
\definecolor{matlab-rood}{rgb}{0.85,0.325,0.098}

\begin{document}
\title{Dynamic Narrowing of VAE Bottlenecks Using GECO and $L_0$ Regularization}

\author{Cedric De Boom,
        Samuel Wauthier,
        Tim Verbelen,
        Bart Dhoedt\\
        \textit{IDLab -- Department of Information Technology, Ghent University -- imec}\\
\textit{Technologiepark-Zwijnaarde 126, 9052 Ghent, Belgium} \\
\textit{firstname.lastname@ugent.be}
}

\maketitle

\begin{abstract}
When designing variational autoencoders (VAEs) or other types of latent space models, the dimensionality of the latent space is typically defined upfront.
In this process, it is possible that the number of dimensions is under- or overprovisioned for the application at hand.
In case the dimensionality is not predefined, this parameter is usually determined using time- and resource-consuming cross-validation.
For these reasons we have developed a technique to shrink the latent space dimensionality of VAEs automatically and on-the-fly during training using Generalized ELBO with Constrained Optimization (GECO) and the $L_0$-Augment-\textsc{Reinforce}-Merge ($L_0$-ARM) gradient estimator.
The GECO optimizer ensures that we are not violating a predefined upper bound on the reconstruction error.
This paper presents the algorithmic details of our method along with experimental results on five different datasets.
We find that our training procedure is stable and that the latent space can be pruned effectively without violating the GECO constraints.
\end{abstract}


% \fi
\IEEEpeerreviewmaketitle

\section{Introduction}
\IEEEPARstart{D}{eep} neural networks are constructed and trained such that intermediate, latent representations of the input data are learned \cite{Goodfellow:2016wc}.
For this reason, deep learning and representation learning are two terms that are closely related.
The goal of representation learning is to project high-dimensional data points such as images, documents, etc.~into a vector space that is typically of low dimensionality compared to the original data points.
In this space, similar data points ideally have similar representations, i.e.~the points lie close to each other in the vector space in terms of some distance metric \cite{Bengio:fm2013,Mikolov:2013wc}.
The low number of dimensions and the similarity property pave the way for applications such as recommender systems (detecting similar items), anomaly detection (detecting contrasting items), data generation, data interpolation, etc.
In these applications it is highly preferred that the vector space dimensionality is kept to a minimum needed for the task at hand.
Not only does it allow for more efficient calculations in the data pipeline, it also lowers the required storage capacity.

Since the arrival of deep learning onto the machine learning scene, we have seen an explosion of models and techniques that are tuned to the task of representation learning and dimensionality reduction.
As opposed to PCA and many other classic algorithms that model linear dependencies, deep neural networks have the powerful ability to learn highly non-linear mappings. %between vectors of data. 
One powerful class of such models are the autoencoders, where the input data is first projected (`encoded') onto a low-dimensional subspace, after which this projection is used to reconstruct the original input data itself (`auto') as well as possible.
Mathematically, if the input of the autoencoder is a $d$-dimensional vector $\bm{x}$, the encoder $f_\text{enc}: \mathbb{R}^d \rightarrow \mathbb{R}^n$ produces a $n$-dimensional projection $\bm{z}$, with $n \ll d$.
% Afterwards, the projection $\bm{z}$ is often processed by an information-limiting operation $h_\text{lim}: \mathbb{R}^n \rightarrow \mathbb{R}^n$, such as a quantizer, sampler or noise inducer.
% We will denote the output of $\limiter{\bm{z}}$ as $\bm{z}'$.
The decoder $g_\text{dec}: \mathbb{R}^n \rightarrow \mathbb{R}^d$ then takes $\bm{z}$ as input and produces an $d$-dimensional output $\bm{x}'$:
\begin{align}
\label{eq:general_autoencoder}
    \bm{z} = \encoder{\bm{x}}; \quad
    \bm{x}' = \decoder{\bm{z}}.
\end{align}
%These equations are general formulations and describe most of the existing autoencoder variants.
From these equations, it becomes clear that autoencoders actually consist of two neural networks, which together create a diabolo architecture: the input and output dimensionality is the same, but the intermediate representation has a lower dimensionality, thereby creating the data `bottleneck'.
The projection $\bm{z}$ has also been called `latent vector', `latent code' or `hidden representation', and `latent space' is often used to denote the vector space of all possible codes.

To make sure the output $\bm{x}'$ resembles the input $\bm{x}$ as closely as possible, the loss function will usually contain a reconstruction error term reflecting either mean squared error (MSE), negative log-likelihood (NLL), a perceptual or adversarial loss, or any other metric or divergence expression (we will elaborate further on loss functions for autoencoders in Section \ref{sec:vaes}).
There are of course many (hyper)parameters that determine the ability of the decoder to reconstruct the original input.
One of these parameters is the width of the bottleneck, i.e.~the dimensionality of the latent vector $\bm{z}$.
The stronger the bottleneck, the more the input data will be compressed by the encoder, thereby potentially causing information loss, which adheres to the information bottleneck (IB) principle as a ``trade-off between compression and prediction'' \cite{Tishby:2015wj,Alemi:2016tb}.
The amount and nature of the compression depends highly on the model architecture and the reconstruction loss.
For example, if we decode an entire image by minimizing a pixel-based $L_2$ error, it is known that the reconstructions will be an increasingly blurred out version of the original image when the bottleneck gets tightened \cite{Pathak:2016vz,Isola:2016tp}, causing the latent vectors to focus on low-frequency information.

Similar to PCA, the bottleneck dimensionality of autoencoders is predefined in the model architecture.
For a given problem statement or use case, one typically provisions an adequate amount of latent dimensions in such a way that the application's requirements are met.
Requirements can be quality-centered, e.g.~accuracy or reconstruction error; performance-based, such as memory consumption or processing time; or application-driven, for example if the latent vectors in existing subsystems already have a fixed dimensionality.
The general conception reads ``the more latent dimensions, the better'', at least in order to improve the quality-based metrics.
This line of reasoning is indeed backed by the information theoretic result that the conditional entropy of a random variable can only decrease for every additional random variable we condition on.
% This line of thought is indeed backed by information theory: consider a random variable $Z_{1:k}$ that represents a valid latent vector $\bm{z}$ with $k$ dimensions, and a random variable $X'$ that stands for the output of the decoder.
% If an additional one-dimensional variable $Z_{k+1}$ is added to the latent vector, and if we treat the decoder as an information channel, it is known that (with $\centropy{\cdot}{\cdot}$ denoting conditional entropy):
% \begin{align}
%     \centropy{X'}{Z_{1:k}} \geq \centropy{X'}{Z_{1:k}, Z_{k+1}}.
% \end{align}
% since for the conditional mutual information we know that
% \begin{align}
%     \cminformation{X'}{Z_{1:k}}{Z_{k+1}} \geq 0.
% \end{align}
More specifically for autoencoders, increasing the latent space with extra dimensions might reduce the uncertainty about the decoder output.
Indeed, it is in general always possible for the decoder to ignore the extra information in the latent vector if it would lead to deteriorated reconstructions.
% Of course, in reality, the new latent space is not just a mere expansion of the original space in a single dimension, but is almost always an entirely different space compared to the original one; the argument above remains, however, valid for illustrative purposes.

In this paper we will try to answer the exact opposite question: to what extent can we eliminate dimensions from the state space with a minimal sacrifice on the output quality?
Or, more explicitly, given a set of conditions that need to be satisfied---e.g.~a maximum classification error---can we effectively prune away latent dimensions without violating these conditions.
This can be achieved by often resource- and time-consuming hyperparameter optimization techniques.
The core of this work, however, is to devise a learning scheme that allows us to dynamically prune the latent space on-the-fly \emph{during} training, without any additional hyperparameter tuning.
% During training, we will make the information bottleneck trade-off between accuracy and compression very explicit by means of a constraint-based formulation of the problem using GECO (Generalized ELBO with Constrained Optimization) \cite{Rezende:wu}, which will be explained in Section \ref{sec:vaes}.
% In this optimization framework, tuned to variational autoencoders (VAEs), the regularization terms will only contribute significantly to the overall loss when all constraints are satisfied.
% The main constraint will be a predefined upper bound on the reconstruction error in this paper, while an $L_0$ regularization term will try to prune the latent dimensions as much as possible.
% The interplay between these two quantities in GECO will have the effect that pruning is encouraged whenever the reconstruction error is below the upper bound; and vice versa, if the quality of the reconstructions becomes too low, pruning is highly discouraged.
% Since we are working with reconstruction losses, but are training without specific target labels, our approach can be deemed a semi-supervised technique for representation learning.

We will show that VAE bottlenecks can effectively be pruned with our methodology.
And even more, we will provide convincing experimental evidence that more dimensions can be eliminated if the constraints are relaxed; that is, if we are satisfied with an overall higher reconstruction error.%, we can potentially achieve an increased prune rate.
The other way around, pushing hard on the reconstruction quality will require more latent dimensions to compress the data, which clearly demonstrates the information bottleneck trade-off.
The remainder of this paper is structured as follows.
First, we will give background information on VAEs and GECO, after which we will provide details on the $L_0$-ARM (Augment-\textsc{Reinforce}-Merge) gradient estimator \cite{Li:2019wg,Yin:2019tm}.
% Next, we will introduce GECO as optimization framework for VAEs in which we will combine the regularization and reconstruction losses.
In Section \ref{sec:methodology} our training procedure will be explained.
Finally, we will conduct a set of experiments to validate our approach on five different datasets in Section \ref{sec:experiments}.

\section{Background Material}
\label{sec:background}
In this section we will cover the theoretical building blocks that are used in the algorithm of Section \ref{sec:methodology}.
We will first cover variational autoencoders and the GECO extension optimization procedure.
Afterwards we will explain the details of the $L_0$-ARM gradient estimator.

\subsection{Variational Autoencoders and GECO}
\label{sec:vaes}
%In traditional autoencoders the latent vectors are deterministic for a given input data point.
In variational autoencoders (VAEs) the encoder models a distribution $q_\theta\!\left( \bm{z} \;\middle\vert\; \bm{x} \right)$ over latent vectors \cite{Kingma:2014tz}.% rather than a single fixed vector .
This distribution is called the approximate posterior and is learned to approximate the true, but unknown posterior distribution $p\!\left( \bm{z} \;\middle\vert\; \bm{x} \right)$.
This is typically done by modeling $q_{\bm{\theta}}$ as a Gaussian distribution with diagonal covariance.
In practice, the neural network $f_\text{enc}^{\bm{\theta}}: \mathbb{R}^d \rightarrow \mathbb{R}^n \times \mathbb{R}^n$ with parameters $\bm{\theta}$ produces both the mean and logarithmic standard deviations of this distribution given some input data $\bm{x}$.
A latent vector $\bm{z}$ is then sampled from the approximate posterior as follows:
\begin{align}
\label{eq:vae_sampling}
    \begin{split}
    \left( \bm{\mu}, \log \bm{\sigma} \right) &= f_\text{enc}^{\bm{\theta}}\!\left(\bm{x}\right);\\
    \bm{z} &\sim q_\theta\!\left( \bm{z} \;\middle\vert\; \bm{x} \right) = \mathcal{N}\!\left( \bm{z}; \bm{\mu}, \mathrm{diag}(\bm{\sigma}) \right).
    \end{split}
\end{align}
The decoder neural network $g_\text{dec}^{\bm{\phi}}: \mathbb{R}^n \rightarrow \mathbb{R}^d$ with parameters $\bm{\phi}$ models the likelihood distribution $p_{\bm{\phi}}\!\left( \bm{x} \;\middle\vert\; \bm{z} \right)$, and produces a reconstructed data point:
\begin{align}
    \bm{x}' = g_\text{dec}^{\bm{\phi}}\!\left(\bm{z}\right).
\end{align}
%In general, by incorporating uncertainty into the model, variational autoencoders enforce the latent space to be semantically coherent, since similar data points will be projected onto close-by regions of the latent space.
Both the encoder and decoder networks are optimized using the negative ELBO \cite{Kingma:2014tz}:
\begin{align}
\label{eq:elbo}
\begin{split}
    \mathcal{L}_{\bm{\theta}, \bm{\phi}}\!\left( \bm{x} \right) &= - \expectationwrt{\bm{z} \sim q_{\bm{\theta}}\!\left( \bm{z} \;\middle\vert\; \bm{x} \right)}{\log p_{\bm{\phi}}\!\left( \bm{x} \;\middle\vert\; \bm{z} \right)}\\
    &\qquad\qquad + \kl{q_{\bm{\theta}}\!\left( \bm{z} \;\middle\vert\; \bm{x} \right)}{p\!\left(\bm{z}\right)}.
\end{split}
\end{align}
The first term in this loss function represents the negative log-likelihood of the data given a latent vector sampled from the posterior distribution.
Differentiability of this term is ensured by the so-called reparameterization trick, in which sampling $\bm{z}$ from the posterior is replaced by sampling a noise vector $\bm{\epsilon}$ from a standard normal distribution and rewriting $\bm{z}$ as a differentiable function of $\bm{\epsilon}$:
\begin{align}
\begin{split}
    \left( \bm{\mu}, \log \bm{\sigma} \right) &= f_\text{enc}^{\bm{\theta}}\!\left(\bm{x}\right);\\
    \bm{\epsilon} &\sim \mathcal{N}\!\left( \bm{z}; \bm{0}, I \right);\\
    \bm{z} &= \bm{\mu} + \bm{\sigma} \odot \bm{\epsilon}.
\end{split}
\end{align}
Here, $\odot$ denotes an element-wise vector product.
In the second term of Equation \eqref{eq:elbo}, $p\!\left(\bm{z}\right)$ is the prior distribution over latents, which is often fixed to a standard normal with zero mean and identity covariance matrix: $p\!\left(\bm{z}\right) = \mathcal{N}\!\left( \bm{z}; \bm{0}, I \right)$.

The KL divergence in Equation \eqref{eq:elbo} can be regarded as a regularization term, as it tries to pull most of the latent factors close to a standard normal distribution.
To make the regularization explicit and tunable, Higgins et al.~proposed $\beta$-VAE in which $\beta$ is the regularization coefficient \cite{Higgins:2017vm}:
\begin{align}
\label{eq:elbo-beta}
\begin{split}
    \mathcal{L}_{\bm{\theta}, \bm{\phi}}\!\left( \bm{x}, \beta \right) &= - \expectationwrt{\bm{z} \sim q_{\bm{\theta}}\!\left( \bm{z} \;\middle\vert\; \bm{x} \right)}{\log p_{\bm{\phi}}\!\left( \bm{x} \;\middle\vert\; \bm{z} \right)}\\
    &\qquad\qquad + \beta\, \kl{q_{\bm{\theta}}\!\left( \bm{z} \;\middle\vert\; \bm{x} \right)}{p\!\left(\bm{z}\right)}.
\end{split}
\end{align}
Increasing $\beta$ puts more weight on the KL term, thereby implicitly tightening the information bottleneck.
Indeed, Higgins et al.~argue that varying and choosing an appropriate value of $\beta$ represents trading off reconstruction quality vs.~latent channel capacity, and that it is generally advised to set $\beta > 1$ in order to arrive at a disentangled latent space \cite{Higgins:2017vm}.

In contrast, Rezende and Viola propose Generalized ELBO with Constrained Optimization (GECO), which rephrases the negative ELBO loss function in terms of a constrained optimization problem using a Lagrange multiplier $\lambda$ \cite{Rezende:wu,Rezende:2018wh}:
\begin{align}
\label{eq:elbo-geco}
\begin{split}
    \mathcal{L}_{\bm{\theta}, \bm{\phi}}\!\left( \bm{x}, \lambda \right) &= 
    \kl{q_{\bm{\theta}}\!\left( \bm{z} \;\middle\vert\; \bm{x} \right)}{p\!\left(\bm{z}\right)}\\
    &\qquad\qquad + \lambda\, \expectationwrt{\bm{z} \sim q_{\bm{\theta}}\!\left( \bm{z} \;\middle\vert\; \bm{x} \right)}{\constraint{\bm{x}, g_\text{dec}^{\bm{\phi}}\!\left(\bm{z}\right)}}.
\end{split}
\end{align}
This Lagrangian optimizes the KL divergence subject to $\expectationwrt{\bm{z}}{\constraint{\bm{x}, g_\text{dec}\!\left(\bm{z}\right)}} \leq 0$, for a given constraint function $\mathcal{C}$ for which $\constraint{\bm{x}, g_\text{dec}\!\left(\bm{z}\right)} \in \mathbb{R}$ (we have left out the parameters for reading comfort)\footnote{The original, more general formulation allows for multiple constraints and Lagrange multipliers, but in this work we will focus on a single constraint.}.
A GECO constraint can essentially be any condition or metric the VAE needs to satisfy, e.g.~an upper bound on a predefined reconstruction ($L_2$) error: %but the constraint will usually model an upper bound on a predefined reconstruction error, e.g.~the $L_2$ error:
\begin{align}
    \label{eq:constraint}
	\constraint{\bm{x}, g_\text{dec}^{\bm{\phi}}\!\left(\bm{z}\right)} = \norm{\bm{x} - g_\text{dec}^{\bm{\phi}}\!\left(\bm{z}\right)}_2 - \tau,
\end{align}
or any other useful divergence.
The hyperparameter $\tau \geq 0$ represents a desired tolerance or upper bound on the reconstruction error, such that $\expectationwrt{\bm{z}}{\constraint{\bm{x}, g_\text{dec}\!\left(\bm{z}\right)}}$ indeed becomes negative if the upper bound is satisfied.
%The value of $\tau$ can be tuned according to the application needs.%, which makes $\tau$ a hyperparameter.
One might argue that tuning $\tau$ is not easier than picking an appropriate $\beta$ in $\beta$-VAE.
However, tuning parameters of latent spaces is an abstract operation, while tuning quality metrics in the data space is much more tangible and practical since its effect can be observed directly in the data reconstruction quality.

Optimizing the Lagrangian in Equation \eqref{eq:elbo-geco} is done through min-max optimization: it is minimized w.r.t.~parameters $\bm{\theta}$ and $\bm{\phi}$, and maximized w.r.t.~$\lambda$.
The details on how to optimize the multiplier $\lambda$ are explained in the original work by Rezende and Viola, and will be further clarified in Section \ref{sec:methodology}.

\subsection{$L_0$ Regularization and ARM}
\label{sec:arm}
There has been considerable effort in the past to sparsify neural networks by pruning redundant or low-magnitude weights \cite{Molchanov:2017wh,Li:2017th,Dai:2018vp}.
Louizos et al.~introduced a learning method that employs $L_0$ regularization \cite{Louizos:2018tl}.
The $L_0$ norm of a numerical vector represents the number of non-zero components, so that minimizing the $L_0$ norm comes down to setting as much vector components to zero as possible.
This is useful for sparsification of neural networks as follows.
Consider the vector $\bm{w}$ of all layer weights in a network, and a binary vector $\bm{\nu}$ (containing only $0$s and $1$s) with the same dimensionality as $\bm{w}$.
By performing the element-wise vector product $\bm{w} \odot \bm{\nu}$, the vector $\bm{\nu}$ acts as an on--off gating mechanism: for every 0 in $\bm{\nu}$ the corresponding weight in the neural network is essentially eliminated from the computational graph.
Therefore, $L_0$ regularization on $\bm{\nu}$ comes down to eliminating weights from the neural network.

Minimizing the $L_0$ norm $\norm{\bm{\nu}}_0$ in a gradient descent context is, however, not straightforward, since it is not differentiable w.r.t.~$\bm{\nu}$.
More specifically, because $\norm{\bm{\nu}}_0$ takes discrete values, it has zero gradients everywhere in its domain.
Gradient estimators such as \textsc{Reinforce} \cite{Williams:1992fi}, Straight-Through estimation \cite{Bengio:2013vv}, and Hard Concrete estimation \cite{Louizos:2018tl} can overcome this issue.
More recently, Augment-\textsc{Reinforce}-Merge or ARM was presented as an unbiased and low-variance gradient estimator that outperforms previously mentioned estimators on a variety of tasks \cite{Yin:2019tm}, and it has been applied successfully to $L_0$-based network sparsification by Li and Ji shortly after its introduction \cite{Li:2019wg}.
It works as follows, given an arbitrary neural network $f$ with $n$ parameters $\bm{w}$, a gating vector $\bm{\nu}$ with dimensionality $n$, and a loss function $\mathcal{L}_{\bm{w}, \bm{\nu}}$ for which:
\begin{align}
\label{eq:l0arm-1}
    \mathcal{L}_{\bm{w}, \bm{\nu}}(\bm{x}, \beta) &= \mathcal{E}\!\left(f(\bm{x}; \bm{w} \odot \bm{\nu}), y\right) + \beta \norm{\bm{\nu}}_0 \nonumber\\
    &= \mathcal{E}\!\left(f(\bm{x}; \bm{w} \odot \bm{\nu}), y\right) + \beta \sum_{j=1}^{n} \mathbbm{1}_{[ \nu_j \neq 0 ]}.
\end{align}
In the above expression, $\mathcal{E}$ represents any error metric between the output of $f$ and a label $y$, $\mathbbm{1}$ is the indicator function, and $\beta$ is a regularizer.
Both terms in Equation \eqref{eq:l0arm-1} are not differentiable w.r.t.~$\bm{\nu}$.
To overcome this issue, we model the components $\nu_j$ as samples from a Bernoulli distribution with parameters $\pi_j$: $\nu_j \sim \mathrm{Ber}(\nu; \pi_j)$.
%We can use this to calculate an upper bound for the loss function:
An upper bound of $\min_{\bm{\nu}} \mathcal{L}_{\bm{w}, \bm{\nu}}$ can then be calculated \cite{Bird:2018uz}:
\begin{align}
\label{eq:l0arm-2}
\begin{split}
    &\min_{\bm{\nu}} \mathcal{L}_{\bm{w}, \bm{\nu}}(\bm{x}, \beta)\\
    &\; \leq \expectationwrt{\bm{\nu} \sim \prod_{j=1}^n \mathrm{Ber}(\nu; \pi_j)}{\mathcal{E}\!\left( f(x; \bm{w} \odot \bm{\nu}), y \right)} + \beta \sum_{j=1}^{n} \pi_j.
\end{split}
\end{align}
We write the right-hand side of \eqref{eq:l0arm-2} as $\hat{\mathcal{L}}_{\bm{w}, \bm{\gamma}}$.
The regularizing term is now differentiable w.r.t.~$\bm{\pi}$, but the first term is still problematic.
We will use ARM to estimate the gradients of this term, after writing the Bernoulli parameters $\bm{\pi}$ as a function of logit parameters $\bm{\gamma}$: $\pi_j = \sigma(\gamma_j)$.
The function $\sigma: \mathbb{R} \rightarrow [0, 1]$ is ideally antithetic and symmetric around its inflection point, e.g.~the sigmoid function: $\sigma(\gamma) = \nicefrac{1}{\left(1 + \exp(-\gamma)\right)}$.
With this reparameterization, and the shorthand notation $\mathcal{F}(\bm{\nu}) = \mathcal{E}\!\left( f(x; \bm{w} \odot \bm{\nu}), y \right)$, Equation \eqref{eq:l0arm-2} can be written as \cite{Yin:2019tm}
\begin{align}
\label{eq:l0arm-3}
\begin{split}
     &\hat{\mathcal{L}}_{\bm{w}, \bm{\gamma}}(\bm{x}, \beta)\\
     &\; = \expectationwrt{\bm{u} \sim \prod_{j=1}^n \!\mathrm{Unif}(u; 0, 1)}{\mathcal{F}(\mathbbm{1}_{[\bm{u} < \sigma(\bm{\gamma})]})} + \beta \sum_{j=1}^{n} \sigma(\gamma_j).
\end{split}
\end{align}
Here, sampling gates from a Bernoulli distribution is replaced by sampling from a uniform distribution between 0 and 1.
The gradient of $\hat{\mathcal{L}}_{\bm{w}, \bm{\gamma}}$ w.r.t.~$\bm{\gamma}$ can now be estimated---unbiased and with low variance---as \cite{Li:2019wg,Yin:2019tm}:
\begin{align}
\label{eq:l0arm-gradient}
\begin{split}
     &\nabla_{\!\bm{\gamma}} \hat{\mathcal{L}}_{\bm{w}, \bm{\gamma}}(\bm{x}, \beta)\\
     &\quad \approx \mathbb{E}_{\bm{u} \sim \prod_{j=1}^n \!\mathrm{Unif}(u; 0, 1)}\!\bigg[\\
     &\qquad
         \left(\mathcal{F}(\mathbbm{1}_{[\bm{u} > \sigma(-\bm{\gamma})]}) - \mathcal{F}(\mathbbm{1}_{[\bm{u} < \sigma(\bm{\gamma})]}) \right) \left( \bm{u} - \frac{1}{2} \right)
     \bigg] \\
     &\qquad + \beta \sum_{j=1}^{n} \nabla_{\!\bm{\gamma}}\sigma(\gamma_j).
\end{split}
\end{align}
This simple estimator has but one disadvantage, which is that two forward passes through the network are required.
% In the original paper, the authors introduce the AR-estimator to overcome the double forward pass, but it leads to a higher variance; we will therefore stick to standard ARM in this work.
We also have to point out that $\sigma$ is smooth and not guaranteed to be exactly zero for closed gates.
However, in line with the original research by Li and Ji, we observe that $\nu_j$ is either close to 0 or close to 1 after training, which means that after sampling the gates are almost always entirely closed or entirely open \cite{Li:2019wg}.
For this reason, at inference time, we decide to explicitly close gates for which $\sigma(\gamma_j) \leq 0.5$, i.e.~$\nu_j$ is set to $0$, such that the corresponding latent dimensions are effectively eliminated.

\paragraph*{Why we need $L_0$ regularization}
With $L_0$ regularization we effectively achieve the desired on-off behavior: gates are completely open or closed.
In the past, $L_1$ and $L_2$ regularization have been used to prune weights and filters from (convolutional) neural networks \cite{Li:2017tha}.
This ensures that the components become smaller, but not exact zero \cite{Li:2019wg}.
And this is needed for the purpose of pruning, since a very tiny but non-zero component is still able to carry information, and is therefore not eliminated effectively.
Put differently, it is possible to make the $L_1$ norm of all gates arbitrarily small since the corresponding weights can be made arbitrarily large. %, and it is therefore not a transparent pruning technique.
 This can be countered by applying additional $L_2$ regularization on the weights themselves, but it would pose an extra optimization trade-off.
Earlier work has therefore often proceeded in a multi-stage fashion: first train the network using $L_1$ regularization, then prune all small weights below a certain threshold, and finally finetune the (smaller) network \cite{Li:2017tha,Liu:2019vv}.
In this work, on the other hand, we only need a single stage in which the network is both trained and pruned at the same time.
Next to this, we do not need an additional threshold hyperparameter to determine the fraction of weights that will be pruned.
Instead we use GECO to make this decision entirely transparent, as will be explained in the next section.

\begin{algorithm}[t!]
 \SetEndCharOfAlgoLine{}  %suppress semicolons
 \SetKwInOut{Input}{Input}
 \SetKwInOut{Parameter}{Parameters }
 \SetKwInOut{Output}{Results}
 \Input{Dataset $\mathcal{D}$}
 \Parameter{$\tau$, $\alpha$, $k$, $\lambda_{\mathrm{min}}$, $\lambda_{\mathrm{max}}$}
 \Output{Learned parameters $\bm{\theta}$, $\bm{\phi}$, $\bm{\gamma}$ and $\lambda$}
 \textit{hit} $\leftarrow$ False\;
 \ForEach{$\bm{x}_i \in \mathcal{D}$ }{
    \eIf{hit}{
        Sample $\bm{u} \sim \prod_{j=1}^n \!\mathrm{Unif}(u; 0, 1)$\;
        $\bm{\nu}_1 \leftarrow \mathbbm{1}_{[\bm{u} > \sigma(-k\bm{\gamma})]}$\;
        $\bm{\nu}_2 \leftarrow \mathbbm{1}_{[\bm{u} < \sigma(k\bm{\gamma})]}$\;
    }{
        $\bm{\nu}_1 \leftarrow \bm{1}$\;\label{algo:line:gates_1}
        $\bm{\nu}_2 \leftarrow \bm{1}$\;\label{algo:line:gates_2}
    }
    Sample $\bm{z} \sim f_\text{enc}^{\bm{\theta}}\!\left( \bm{x}_i\right)$ with reparameterization trick\;
    \label{algo:line:sample_z}
    $\mathcal{E}_1 \leftarrow \norm{g_\text{dec}^{\bm{\phi}}\!\left( \bm{z} \odot \bm{\nu}_1  \right) - \bm{x}_i}_2 $\;\label{algo:line:reconstruction_1}
    $\mathcal{E}_2 \leftarrow \norm{g_\text{dec}^{\bm{\phi}}\!\left( \bm{z} \odot \bm{\nu}_2  \right) - \bm{x}_i}_2 $\;\label{algo:line:reconstruction_2}
    
    $\mathcal{C} \leftarrow \mathcal{E}_2 - \tau$\;\label{algo:line:geco_constraint}
    \eIf{$i == 0$}{
        $\mathcal{C}_{\mathrm{ma}} \leftarrow \mathcal{C}$\;
    }{
        $\mathcal{C}_{\mathrm{ma}} \leftarrow \alpha \cdot \mathcal{C}_{\mathrm{ma}} + (1 - \alpha) \cdot \mathcal{C}$\;
    }
    $\mathcal{C}' \leftarrow \mathcal{C} + \mathrm{StopGradient}(\mathcal{C}_{\mathrm{ma}} - \mathcal{C})$\;
    $\lambda' \leftarrow \max(\min(\mathrm{softplus}(\lambda)^2, \lambda_{\mathrm{max}}), \lambda_{\mathrm{min}})$\;
    
    $\mathcal{L} \leftarrow \lambda' \cdot \mathcal{C}' + \frac{1}{\norm{\bm{\nu}_2}_0} \norm{\klelem{f_\text{enc}^{\bm{\theta}}\!\left( \bm{x}_i\right)}{p\!\left(\bm{z}\right) } \odot \bm{\nu}_2}_1 $\;\label{algo:line:geco_kl_elem}
    
    \If{$\mathcal{C} \leq 0$}{
        $\mathcal{L} \leftarrow \mathcal{L} + \sum_j \sigma(k\gamma_j)$\;\label{algo:line:geco_l0}
        \textit{hit} $\leftarrow$ True
    }\label{algo:line:geco_end}
    
    $\nabla_{\bm{\phi}, \bm{\theta}, \bm{\gamma}, \lambda}\mathcal{L} \leftarrow \mathrm{Backpropagation}(\mathcal{L})$\;
    \If{hit}{
        $\nabla_{\bm{\gamma}}\mathcal{L} \leftarrow \nabla_{\bm{\gamma}}\mathcal{L} + \lambda' \cdot k \cdot (\mathcal{E}_1 - \mathcal{E}_2)\cdot \left(\bm{u} - \frac{1}{2}\right)$\;\label{algo:line:arm_update}
    }
    Optimize parameters $\bm{\phi}$, $\bm{\theta}$, $\bm{\gamma}$ and $\lambda$ with Adam\;
 }
 \caption{Optimization procedure to shrink the bottleneck dimensionality of a VAE using GECO and $L_0$-ARM gradient estimation.}
 \label{algo:methodology}
\end{algorithm}

\section{Shrinking VAE bottlenecks}
\label{sec:methodology}
In this section we will describe how to combine GECO optimization and the $L_0$-ARM gradient estimator in order to narrow VAE bottlenecks.
% To lift a tip of the veil, the core idea is we will impose constraints on the data reconstruction quality of the VAE while maintaining an $L_0$ regularizer which task it is to prune the latent space as well as possible without violating the constraints.
% For this idea to work, on the one hand we will use GECO to coordinate the constrained optimization, while on the other hand $L_0$-ARM will be used to estimate gradients for the $L_0$ regularizer.
The result of our methodology will be a novel VAE optimization scheme that effectively learns to project data points in a latent space while \textit{at the same time} reducing that latent space's dimensionality and ensuring high-quality reconstructions.
Before diving into the details, we first take a look at a couple of issues with the original $L_0$-ARM weight pruning technique.
First, as Li and Ji state, sparsity strength---i.e.~a measure for the number of weights that can be deleted---needs to be tuned explicitly by the regularization parameter $\beta$ \cite{Li:2019wg}.
Apart from this regularization parameter, gates are encouraged to close themselves deliberately without any explicitly imposed constraints.
Even more, in the original text the authors show that most of the gates are already settled to open or closed after ca.~50 training epochs.
That is, often well before convergence of the main metric, i.e.~the classification or regression error,  % that you care about most for your application needs. 
and this causes a ``drift'' between pruning and performance optimization. %, there are two other concerns regarding the issues stated above. 
Another concern is that the prune rate highly depends on the value of the regularization parameter, and setting this parameter precisely according to one's needs is not transparent and requires hyperparameter optimization. %multiple training phases with different parameter values---i.e.~hyperparameter optimization. 
Finally, since the regularization parameter is our only proxy to tune the prune rate, we have little control over the final performance of the network.
For example, if we are satisfied with a slightly lower predictive accuracy, we can decide to increase the prune rate.
This operation will indeed come at a performance cost, but it is unclear by how much the performance will deteriorate.
Therefore, we have to resort to a complete hyperparameter sweep.
With GECO, however, it is possible to set an upper bound on the reconstruction error during training.
So, instead of tuning the prune rate implicitly with a regularization parameter, we will \textit{explicitly} set a maximum on the reconstruction error and prune as many dimensions as possible without violating this maximum.

Our procedure, as shown in Algorithm \ref{algo:methodology}, is simple yet effective at shrinking the bottleneck dimensionality of VAEs.
At the core is a basic VAE architecture consisting of an encoder $f_\text{enc}^{\bm{\theta}}$ and decoder $g_\text{dec}^{\bm{\phi}}$ neural network, as described in Section \ref{sec:vaes}, and a global gating vector $\bm{\nu}$ as explained in Section \ref{sec:arm}.
The vector $\bm{\nu}$ has the same number of dimensions as the VAE latent space.
For a given data point $\bm{x}$, we determine its latent vector $\bm{z}$ by sampling from the output of the encoder using the reparameterization trick: $\bm{z} \sim f_\text{enc}^{\bm{\theta}}\!\left( \bm{x}\right)$ (line \ref{algo:line:sample_z} in Algorithm \ref{algo:methodology}).
Next, the sampled vector $\bm{z}$ is gated through vector $\bm{\nu}$ using the operation $\bm{z} \odot \bm{\nu}$, thereby deactivating some of the components.
The resulting vector is used by the decoder to reconstruct the original data point: $\bm{x}' = g_\text{dec}^{\bm{\phi}}\!\left( \bm{z} \odot \bm{\nu} \right)$.
To measure the reconstruction quality, we have chosen to use the squared error between the original and reconstructed data points, as shown in lines \ref{algo:line:reconstruction_1} and \ref{algo:line:reconstruction_2}.
By analogy with Equation \eqref{eq:constraint}, we can now define an upper-bound constraint as the difference between the reconstruction error and a pre-defined threshold $\tau$ (line \ref{algo:line:geco_constraint}).
At this point we are able to define the constrained optimization problem that we are solving, in which the Lagrangian incorporates an additional $L_0$ regularization term on the gates:
\begin{align}
\label{eq:elbo-geco-l0}
\begin{split}
    \mathcal{L}_{\bm{\theta}, \bm{\phi}, \bm{\nu}}\!\left( \bm{x}, \lambda \right) &= \kl{f_\text{enc}^{\bm{\theta}}\!\left( \bm{x} \right)}{p\!\left(\bm{z}\right)}
    + \norm{\bm{\nu}}_0\\
    &\qquad\quad + \lambda\, \expectationwrt{\bm{z}}{\constraint{\bm{x}, g_\text{dec}^{\bm{\phi}}\!\left(\bm{z} \odot \bm{\nu} \right)}}. %\nonumber \\
    %\mathrm{s.t.}\quad &\expectationwrt{\bm{z}'}{\constraint{\bm{x}, g_\text{dec}^{\bm{\phi}}\!\left(\bm{z}' \odot \bm{\nu} \right)}} \leq 0.
\end{split}
\end{align}

Lines \ref{algo:line:geco_constraint} to \ref{algo:line:geco_end} in Algorithm \ref{algo:methodology} comprise the GECO optimization details, as implemented by Rezende and Viola \cite{Rezende:2018wh}.
In line \ref{algo:line:geco_kl_elem} we calculate both the constraint term and the KL regularization.
For the latter we have introduced the notation $\mathrm{KL}_{\mathrm{elem}}$ to denote the vector of element-wise KL components.
More specifically, given that $\left( \bm{\mu}, \log \bm{\sigma} \right) = f_\text{enc}^{\bm{\theta}}\!\left(\bm{x}\right)$, parameterizing a multidimensional Gaussian distribution with mean $\bm{\mu}$ and diagonal covariance $\mathrm{diag}(\bm{\sigma})$ as explained in Equation \eqref{eq:vae_sampling}, and $p\!\left(\bm{z}\right)$ a standard normal distribution, the $j$'th component of the element-wise KL is calculated as:
\begin{align}
\begin{split}
    &\klelem{f_\text{enc}^{\bm{\theta}}\!\left( \bm{x}\right)}{p\!\left(\bm{z}\right) }_j \\
    &\qquad = -\frac{1}{2}\left(1+ \log \bm{\sigma}_j - \bm{\sigma}_j - \bm{\mu}_j^2 \right).
\end{split}
\end{align}
Since the right-hand side is always positive, line \ref{algo:line:geco_kl_elem} is just a compact way of writing the average KL divergence, thereby taking into account the pruned components.
After all, once a dimension is eliminated, its KL divergence should not weigh on the optimization loss anymore.
Regarding the constraint term, we entertain a moving average which simulates the expectation of the constraint w.r.t.~$\bm{z}$, as defined in Equation \eqref{eq:elbo-geco-l0}.
The parameter $\alpha$ is set to control the rate of the moving average, with a value typically just slightly lower than $1$.
For efficiency reasons, we stop the gradients from flowing through this moving average, i.e.~only the current constraint (at step $i$) is involved in backpropagation.
We also perform a monotonous squared softplus operation on $\lambda$ to make sure that the multiplier is positive and is able to `move fast', i.e.~to decrease quickly once the constraint is met, and vice versa. 
%This is in line with Deepmind's Sonnet library\footnote{\url{https://github.com/deepmind/sonnet}}. 
We also clamp the values of $\lambda'$ such that they cannot become arbitrarily small or large.
This is needed to prevent $\lambda'$ from growing exponentially.
In line \ref{algo:line:geco_l0}, we add the $L_0$ regularization term to the overall loss when the constraint is satisfied for the current data point, i.e.~we only allow gates to close themselves when we have some wiggle room and are not violating constraints.
This is needed to stabilize the optimization procedure: if we are in the process of eliminating a dimension by gradually decreasing the corresponding $\bm{\gamma}$ component, the number one priority is maintaining the required reconstruction loss.
If this loss would increase too much, we should not proceed, but instead focus on improving the model parameters.

The remaining lines of Algorithm \ref{algo:methodology} constitute the $L_0$-ARM gradient estimation of the gating vector.
First, notice the \emph{hit} variable which is set to True once the constraint is satisfied.
Therefore, at the start of training, the gates are set wide open, and only when the constraints are achieved for the first time, we will start the $L_0$-ARM optimization.
The parameter $k$ in lines \ref{algo:line:gates_1} and \ref{algo:line:gates_2} is introduced in order to scale the sigmoid function; a typical value of $k > 1$ will make the sigmoid shape steeper, thereby allowing faster transitions between open and closed gates.
Most operations are self-explanatory and are in line with Equation \eqref{eq:l0arm-gradient}.
We do want to point out that we only need to sample a single latent vector from the encoder, but that we need two forward passes through the decoder in lines \ref{algo:line:reconstruction_1} and \ref{algo:line:reconstruction_2}.
The resulting reconstruction errors are then used to update the gradients for $\bm{\gamma}$ in line \ref{algo:line:arm_update}.
In this equation, according to the chain rule, we need to multiply by $k$ and by $\lambda'$.
Finally, in the last line, all parameters are optimized using Adam---or any other gradient descent flavor.

\section{Experiments}
\label{sec:experiments}
We will evaluate our methodology in a variety of settings on five different, standard image datasets.
We consider five image datasets, of which the first three are typically used to evaluate VAE performance and latent space disentanglement.
The last two datasets contain realistic images, are not artificially constructed, and have been extensively used in the past to benchmark different varieties of machine learning models.
\begin{itemize}
    \item dSprites (Shapes2D) \cite{dsprites17}: $64\times 64$ images of white 2D shapes on a black background. There are five degrees of freedom: shape (square, ellipse or heart), scale, rotation, $x$-position, $y$-position.
    \item Shapes3D \cite{3dshapes18}: $64\times 64$ images of 3D shapes in a 3D environment. There are six degrees of freedom: shape, scale, rotation, floor hue, wall hue, object hue.
    \item Cars3D \cite{Reed:2015uw}: $64\times 64$ images of 3D car models on a white background. There are three degrees of freedom: object type, elevation, azimuth.
    \item MNIST: $28\times 28$ images of handwritten digits.
    \item CIFAR-100: $32\times 32$ images of real-life pictures of objects; we upscale the images to $64\times 64$ pixels.
\end{itemize}

% architectures
We use a standard convolutional neural network as architecture for the encoder $f_\text{enc}^{\bm{\theta}}$.
For input color images of $64\times 64\times 3$ pixels, we use four convolutional layers with stride 2 and kernel sizes of $4\times 4$ pixels for the first layer and $3\times 3$ pixels for the subsequent layers.
Each of these layers is fed through a leaky ReLU activation with leakiness $0.02$ and a BatchNorm layer with momentum $0.8$.
The output of the final convolutional layer is flattened and used as input for a linear layer with 128 output neurons.
A final linear layer is applied which calculates both a mean and variance prediction; this layer therefore contains $2\cdot n$ output neurons, with $n$ the dimensionality of the latent space.
For MNIST images of $28\times 28\times 1$ pixels, there are only three convolutional layers with resp.~$4\times 4$, $3\times 3$ and $2\times 2$ filters.
The decoder architecture is a mirrored version of the encoder network.
Two linear layers transform the latent vector into the input of four consecutive convolutional layers.
Each of the convolutions has a $3\times 3$ kernel with stride 1, and is preceded by a BatchNorm layer and an upsampling operation which increases the image size by 2 through nearest neighbor interpolation.
Between each layer we apply a leaky ReLU function, and the final activation is a regular sigmoid to obtain pixel values between 0 and 1.
For MNIST we have again three convolutional layers with resp.~kernel sizes $2\times 2$ and two times $3\times 3$.

In all experiments $k=7$, $\alpha=0.99$, $\lambda_{\mathrm{min}} = 10^{-5}$, $\lambda_{\mathrm{max}} = 5$, batch size is 64, learning rate is $10^{-3}$, and we use Adam as optimizer.
We initialize $\lambda = 1$ and all components of $\bm{\gamma}$ are initialized to $0.42$ such that $\sigma(k\cdot 0.42) \approx 0.95$, which means that a gate is open with probability $95\%$.
This is done to ensure training stability.

\subsection{Visualizing the training procedure}

\begin{figure}[t]
    \centering
    %%%%%%%%%%%
%%  MSE  %%
%%%%%%%%%%%
\begin{minipage}[b]{0.495\textwidth}
\centering
\begin{tikzpicture}
\begin{axis}[
    xlabel={Batch number $\cdot 10^3$},
    ylabel={MSE},
    xmin=0, xmax=500,
    ymin=0, ymax=70,
    xtick={0,100,200,300,400,500},
    xticklabels={0,10,20,30,40,50},
    ytick={0,10,20,30,40,50,60,70},
    label style={font=\scriptsize},
    tick label style={font=\scriptsize},
    legend pos=north west,
    ymajorgrids=false,
    grid style=dashed,
    width=\linewidth,height=4cm
]
\addplot+[color=ugent-blauw,
    mark=none]
    table [
    x=Step,
    y=Value,
    col sep=comma,
    each nth point={2}
    ] {data/out_train_ae.csv};
\addplot[
    color=matlab-rood,
    mark=none
    ]
    coordinates {
    (0,20)(500,20)
    };
\end{axis}

\node[inner sep=0pt] (t) at (7.65,0.70)
    {{\color{matlab-rood} $\tau$}};

% Set margins by adjusting the bounding box:
\pgfresetboundingbox
\path
  (current axis.south west) -- ++(-0.25in,-0.3in)
  rectangle (current axis.north east) -- ++(0.07in,0.05in);
\end{tikzpicture}
\end{minipage}\\[0.25cm]
%%%%%%%%%%%%%
%%  GATES  %%
%%%%%%%%%%%%%
\begin{minipage}[b]{0.495\textwidth}
\centering
\begin{tikzpicture}
\begin{axis}[
    xlabel={Batch number $\cdot 10^3$},
    ylabel={$\norm{\sigma(k\bm{\gamma})}_1$},
    xmin=0, xmax=500,
    ymin=0, ymax=10,
    xtick={0,100,200,300,400,500},
    xticklabels={0,10,20,30,40,50},
    ytick={0,2,4,6,8,10},
    label style={font=\scriptsize},
    tick label style={font=\scriptsize},
    legend pos=north west,
    ymajorgrids=false,
    grid style=dashed,
    width=\linewidth,height=4cm
]
\addplot+[color=ugent-blauw,
    mark=none]
    table [
    x=Step,
    y=Value,
    col sep=comma,
    each nth point={2}
    ] {data/out_train_reg.csv};
\end{axis}

% Set margins by adjusting the bounding box:
\pgfresetboundingbox
\path
  (current axis.south west) -- ++(-0.25in,-0.3in)
  rectangle (current axis.north east) -- ++(0.07in,0.05in);

\end{tikzpicture}
\end{minipage}\\[0.25cm]
%%%%%%%%%%%%%%
%%  LAMBDA  %%
%%%%%%%%%%%%%%
\begin{minipage}[b]{0.495\textwidth}
\centering
\begin{tikzpicture}
\begin{semilogyaxis}[
    xlabel={Batch number $\cdot 10^3$},
    ylabel={$\lambda$},
    xmin=0, xmax=500,
    ymin=0, ymax=4,
    xtick={0,100,200,300,400,500},
    xticklabels={0,10,20,30,40,50},
    ytick={0.001, 0.01, 0.1, 1},
    yticklabels={$10^{-3}$,$10^{-2}$,$10^{-1}$,$1$},
    ylabel shift=-3 pt,
    label style={font=\scriptsize},
    tick label style={font=\scriptsize},
    legend pos=north west,
    ymajorgrids=false,
    grid style=dashed,
    width=\linewidth,height=4cm
]
\addplot+[color=ugent-blauw,
    mark=none]
    table [
    x=Step,
    y=Value,
    col sep=comma,
    each nth point={2}
    ] {data/out_train_lambda.csv};
\end{semilogyaxis}

% Set margins by adjusting the bounding box:
\pgfresetboundingbox
\path
  (current axis.south west) -- ++(-0.2in,-0.3in)
  rectangle (current axis.north east) -- ++(0in,0.05in);

\end{tikzpicture}
\end{minipage}\\[0.25cm]
%%%%%%%%%%%%
%%  HITS  %%
%%%%%%%%%%%%
\begin{minipage}[b]{0.495\textwidth}
\centering
\begin{tikzpicture}
\begin{axis}[
    xlabel={Batch number $\cdot 10^3$},
    ylabel={\% tolerance hits},
    xmin=0, xmax=500,
    ymin=0, ymax=1,
    xtick={0,100,200,300,400,500},
    xticklabels={0,10,20,30,40,50},
    ytick={0,0.5,1},
    yticklabels={0,50,100},
    label style={font=\scriptsize},
    tick label style={font=\scriptsize},
    legend pos=north west,
    ymajorgrids=false,
    grid style=dashed,
    width=\linewidth,height=4cm
]
\addplot+[color=ugent-blauw,
    mark=none]
    table [
    x=Step,
    y=Value,
    col sep=comma,
    each nth point={2}
    ] {data/out_train_hits.csv};
\end{axis}

% Set margins by adjusting the bounding box:
\pgfresetboundingbox
\path
  (current axis.south west) -- ++(-0.25in,-0.3in)
  rectangle (current axis.north east) -- ++(0.07in,0.05in);

\end{tikzpicture}
\end{minipage}
    \caption{Overview of four quantities during a typical training run: MSE with the GECO threshold $\tau$ indicated by a red line, the $L_1$ norm of the continuous gate vector (i.e.~the sum of vector components), the Lagrange multiplier $\lambda$ and a measure for the percentage of GECO tolerance hits within a batch.}
    \label{fig:performance_plots}
\end{figure}
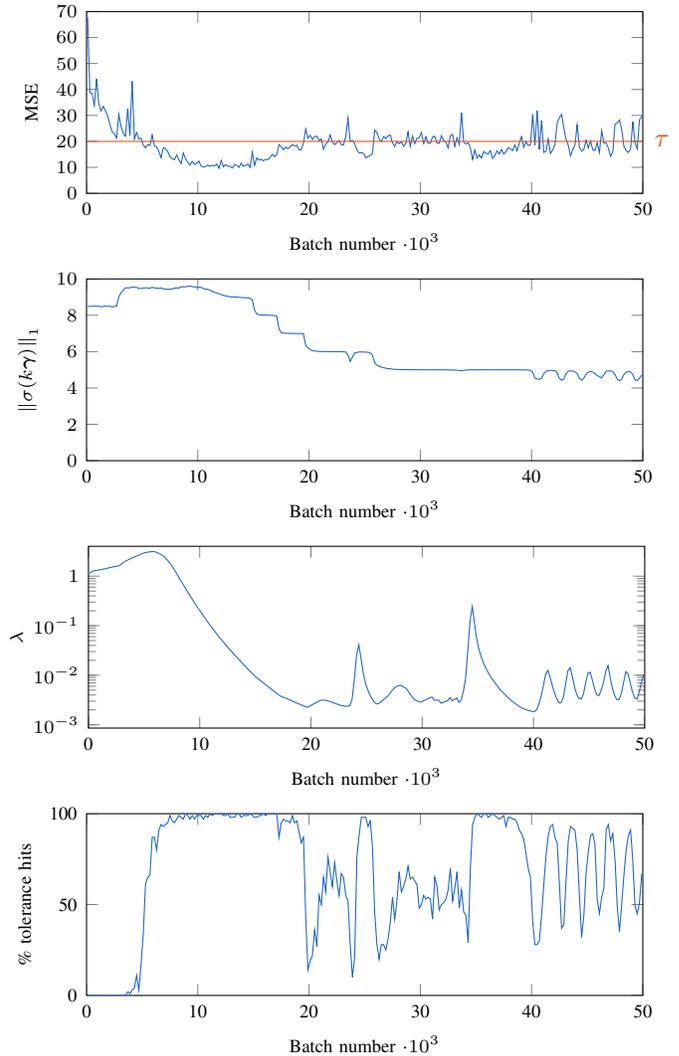

In a first experiment we will observe the training behavior by observing four different quantities over time: the mean squared reconstruction error (MSE), the number of gates that are open or closed, the Lagrange multiplier $\lambda$, and the fraction of data points within a batch that satisfy the constraint.
These graphs are shown in Figure \ref{fig:performance_plots} in which we have trained a VAE on dSprites for 50K batches, with $n=10$, i.e.~the original---and therefore maximum---dimensionality of the VAE.
At the start of training we see that $\lambda$ first increases until the MSE dives below the threshold $\tau$, after which it drops to a small value.
This has the effect that the regularization terms gain more importance, and after ca.~10K batches we notice that the gates are gradually starting to close themselves, which is visible in the descending staircase pattern in the second plot.
This plot shows the $L_1$ norm of the smoothed gate vector $\sigma(k\bm{\gamma})$, which is essentially the sum of the vector components. % since these are all positive.
 The $y$-axis of the graph therefore represents a continuous measure of the number of open gates.
Each plateau in the graph corresponds to an integer amount of opened gates, so we see that after 10K batches there are 9 open gates, which drops to 6 gates at 20K batches, and to 5 gates at 25K batches.
Of course, when more gates are closed, the reconstruction errors are expected to increase, which can be observed in the first graph, and the GECO tolerance hit/miss ratio will drop.
Whenever the MSE is above $\tau$, the multiplier $\lambda$ will increase in order to temporarily attribute more weight to the reconstruction error.
We can visually distinguish two such increases around 25K and 35K batches in the third plot.
As a final remark, we want to bring to attention the oscillatory behavior of the gates after 40K batches.
These oscillations come from the effect that in the process of closing an additional gate, the constraint is violated severely, which causes $\lambda$ to increase in order to lower the MSE in favor of the regularization terms, thereby opening the gate again.
In other words, the model is constantly trying to close a gate, but it is obstructed by the GECO mechanism not allowing any heavy constraint violations.

\begin{figure*}[t]
\centering
\includegraphics[width=.9\textwidth]{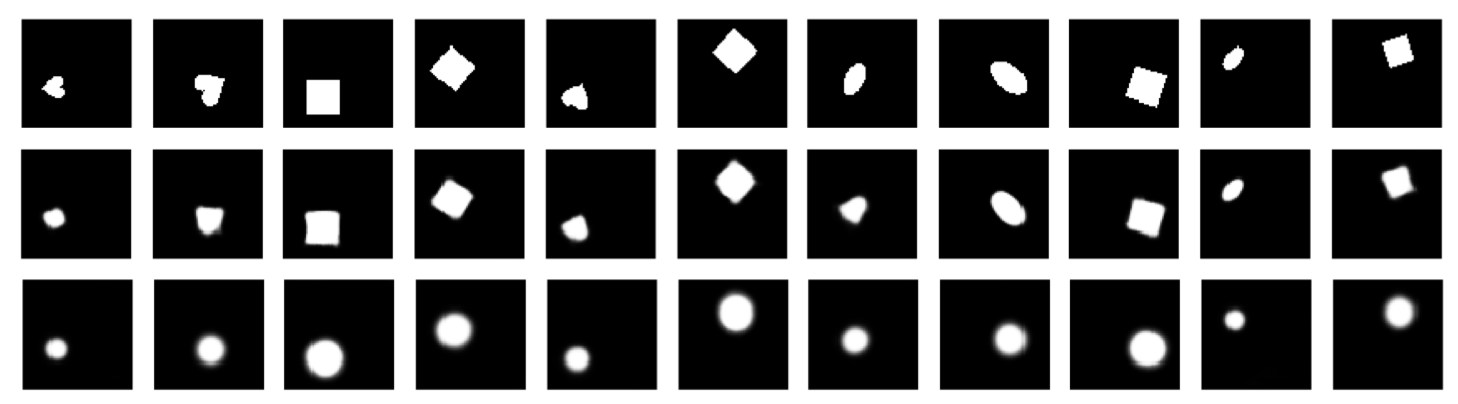}
\caption{Original dSprites images (top row) are reconstructed by a VAE trained with $\tau = 25$ resulting in 5 dimensions (middle row), and $\tau = 35$ resulting in 3 dimensions (bottom row).}
\label{fig:dsprites_complete}
\end{figure*}

\begin{figure}[t]
\centering
\input{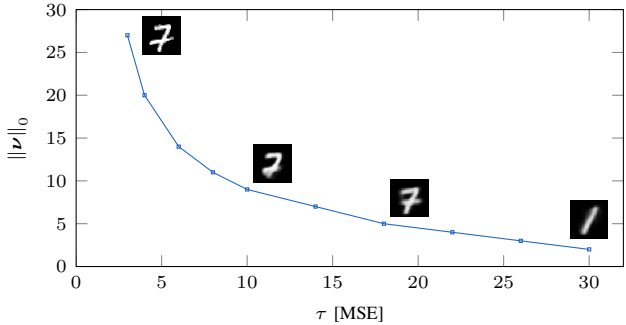}
\caption{This plot represents the information bottleneck pareto front for a convolutional VAE trained on MNIST. It shows the number of retained dimensions in the latent space $\norm{\bm{\nu}}_0$ as a function of the GECO threshold $\tau$.}
\label{fig:bottleneck_front}
\end{figure}

\subsection{Visualizing the information bottleneck}
We will now vary the GECO tolerance and observe the final bottleneck dimensionality after training.
This allows us to test whether an increased tolerance will lead to a tighter bottleneck and vice versa.
We perform tests on the MNIST dataset, for which we set the initial dimensionality $n=32$.
The GECO tolerance is set to different MSEs of 3, 4, 6, 8, 10, 14, 18, 22, 26 and 30, and train for 2000 epochs.
The resulting bottleneck pareto front is shown in Figure \ref{fig:bottleneck_front}, and we can clearly observe the inverse relationship between the threshold $\tau$ and the number of retained latent dimensions $\norm{\bm{\nu}}_0$.
For $\tau = 3$ the model learns to withhold 27 out of 32 dimensions, while for $\tau = 30$ we only need 2 dimensions.
For this specific dataset, neural architecture and training procedure, Figure \ref{fig:bottleneck_front} makes the information bottleneck pareto front very visual and tangible, since it shows to what extent it is possible to tighten the bottleneck for a given GECO threshold.
Next to the plotted bottleneck line, we also show reconstructions of a given image of the number 7 for a few selected points on the graph.
We can clearly see that the quality of the reconstructions deteriorates as we increase $\tau$---for $\tau=30$ the construction resembles a 1 rather than a 7---and that they also become blurrier.
Based on visual samples, it is relatively easy to select the GECO threshold one is comfortable with for the application at hand.

\subsection{Comparing reconstruction qualities}
In the next experiment we train two VAEs on the dSprites dataset with different settings for the GECO threshold: $\tau=20$ and $\tau=35$ (in MSE).
For $\tau=20$ this results in 5 latent dimensions, while $\tau=35$ gives 3 dimensions.
As mentioned before, the dSprites dataset is constructed with 5 degrees of freedom (shape, scale, rotation, $x$- and $y$-position).
In Figure \ref{fig:dsprites_complete} we can observe that in the middle row all five variables are used to reconstruct the original image.
In the bottom row, however, we only see blurred circles at the positions where the original shapes are located.
This means that both shape and rotational information is discarded, which indeed corresponds to the 3 retained latent dimensions instead of 5.
While this line of reasoning may sound logical, it is far from a rigorous proof that the 3 dimensions indeed each correspond with the remaining degrees of freedom (scale, $x$- and $y$-position).
Since we are not explicitly striving for disentangled latent spaces in this work, we will not go deeper into this matter.

\begin{table*}[t]
    %\scriptsize
    \setlength\tabcolsep{2mm}
    \def\arraystretch{1.1}
    \centering
    \begin{tabular}{llllll}
    \toprule
    & $\bm{n}$ & \textbf{MSE (200K)} & $\norm{\bm{\nu}}_{\bm{0}}$ \textbf{(200K)} & $\norm{\bm{\nu}}_{\bm{0}}$ \textbf{(300K)} & $\norm{\bm{\nu}}_{\bm{0}}$ \textbf{(400K)} \\\hline
    dSprites & 5 & 9.48 & 7 & 6 & 6 \\
    & 10 & 8.83 & 7 & 6 & 6 \\
    & 30 & 6.14 & 13 & 12 & 10 \\\hline
    Cars3D & 5 & 94.42 & 6 & 5 & 5 \\
    & 10 & 65.67 & 14 & 13 & 12 \\
    & 30 & 60.97 & 21 & 17 & 16 \\\hline
    Shapes3D & 5 & 36.73 & 6 & 6 & 6 \\
    & 10 & 19.53 & 9 & 9 & 9 \\
    & 30 & 18.74 & 20 & 15 & 11 \\\hline
    MNIST & 5 & 17.07 & 5 & 5 & 5 \\
    & 10 & 9.49 & 10 & 9 & 9 \\
    & 30 & 7.16 & 14 & 14 & 14 \\\hline
    CIFAR-100 & 5 & 325.87 & 6 & 6 & 6 \\
    & 10 & 249.1 & 10 & 10 & 10 \\
    & 30 & 147.8 & 31 & 30 & 30 \\
    \bottomrule
    \end{tabular}
    \caption{For five different datasets, we train a regular VAE with bottleneck dimensionality $n$ and GECO threshold $\tau=0$. The obtained MSE is shown after a training budget of 200K batches. We train VAEs with $L_0$ regularization and initial bottleneck dimensionality $2n$, we set the GECO threshold $\tau$ to the previously obtained MSE, and the resulting 0-norm of the gating vector after 200K, 300K and 400K batches is shown.}
    \label{tab:compression}
\end{table*}

\subsection{Evaluating compression efficiency}
In this final experiment we want to find out how efficient our method is at pruning latent dimensions.
For this purpose, we devise an approach in which we first train a baseline regular VAE with a predefined number of latent dimensions $n$ and with a GECO tolerance of 0 MSE, which means that we want to squeeze out every bit of reconstruction quality there is to gain, thereby sacrificing KL regularization.
We give the model a fixed training budget of 200K batches, after which we write down the moving average (with factor $0.95$) of the final reconstruction error.
Then, in a second step, we train a new VAE with $L_0$ regularization and GECO that we provision with $2n$ latent dimensions, and we observe to what extent this number can be lowered to $n$ given the obtained reconstruction error in step one as GECO tolerance.
We do this for all five datasets, for values $n=5, 10, 30$, and for fixed training budgets of 200K, 300K and 400K batches.
This will allow us to measure the efficiency of our approach in terms of how fast we can reduce the latent dimensionality compared to the original VAE.
The resulting numbers are presented in Table \ref{tab:compression}.
For most configurations, our approach is able to prune the bottleneck very effectively, coming close to the original value of $n$, sometimes achieving the same value or even surpassing it.
This is often true for $n=30$ (except for CIFAR-100), which shows that the original VAE was probably overprovisioned.
We also observe that we are coming close to the optimal number of dimensions already after a training budget of 200K batches, i.e.~the same budget with which we have trained the original VAE.
This is probably thanks to the higher initial dimensionality of $2n$, which allows the reconstruction error to drop more quickly below the GECO treshold at the start of training, after which we can start pruning the bottleneck.
Only in a select number of cases---again, mostly for $n=30$---are we able to further reduce the dimensionality significantly if we continue training for 400K batches.
In summary, Table \ref{tab:compression} shows that our method is widely applicable in different settings and that it can provide near-optimal dimensionality reductions without incurring a large additional training budget.

\section{Conclusion}
We have devised an algorithm to reduce the bottleneck dimensionality of VAEs on-the-fly during training.
For this purpose we have used $L_0$ regularization and the $L_0$-ARM gradient estimator to train a gating mechanism that either blocks or passes information for each latent factor independently.
To decide how many factors are needed, we employ GECO to define constraints as upper bounds on the reconstruction error.
In the experiments we show that our algorithm is effective at reducing the latent dimensionality on five different datasets.
It is also a useful tool to assess whether a VAE bottleneck is over- or underprovisioned.
The only downside of the algorithm is the (small) computational overhead that comes from a second forward pass through the decoder network, as required by $L_0$-ARM.
In future work one can look at applying the method to all intermediate representations \cite{Dai:2018vp}, and not only to the encoder output.
It would also be interesting to see if it can be applied to recommender systems or VAE-based models for control tasks, e.g.~in robotics.

\section*{Acknowledgments}
This research received funding from the Flemish Government under the ``Onderzoeksprogramma Artifici\"ele Intelligentie (AI) Vlaanderen'' programme.
\ifCLASSOPTIONcaptionsoff
  \newpage
\fi



%\bibliographystyle{IEEEtran}

\bibliographystyle{IEEEtran}
\bibliography{lib}

\begin{thebibliography}{10}
\providecommand{\url}[1]{#1}
\csname url@samestyle\endcsname
\providecommand{\newblock}{\relax}
\providecommand{\bibinfo}[2]{#2}
\providecommand{\BIBentrySTDinterwordspacing}{\spaceskip=0pt\relax}
\providecommand{\BIBentryALTinterwordstretchfactor}{4}
\providecommand{\BIBentryALTinterwordspacing}{\spaceskip=\fontdimen2\font plus
\BIBentryALTinterwordstretchfactor\fontdimen3\font minus
  \fontdimen4\font\relax}
\providecommand{\BIBforeignlanguage}[2]{{%
\expandafter\ifx\csname l@#1\endcsname\relax
\typeout{** WARNING: IEEEtran.bst: No hyphenation pattern has been}%
\typeout{** loaded for the language `#1'. Using the pattern for}%
\typeout{** the default language instead.}%
\else
\language=\csname l@#1\endcsname
\fi
#2}}
\providecommand{\BIBdecl}{\relax}
\BIBdecl

\bibitem{Goodfellow:2016wc}
I.~Goodfellow, Y.~Bengio, and A.~Courville, \emph{{Deep Learning}}.\hskip 1em
  plus 0.5em minus 0.4em\relax MIT Press, 2016.

\bibitem{Bengio:fm2013}
Y.~Bengio, A.~Courville, P.~V. I. t.~o. pattern, and {2013}, ``{Representation
  learning: A review and new perspectives},'' \emph{IEEE Transactions on
  Pattern Analysis and Machine Intelligence}, 2013.

\bibitem{Mikolov:2013wc}
T.~Mikolov, K.~Chen, G.~Corrado, and J.~Dean, ``{Efficient Estimation of Word
  Representations in Vector Space},'' in \emph{Proceedings of Workshop at
  ICLR}, 2013.

\bibitem{Tishby:2015wj}
N.~Tishby and N.~Zaslavsky, ``{Deep Learning and the Information Bottleneck
  Principle},'' \emph{arXiv.org}, 2015.

\bibitem{Alemi:2016tb}
A.~A. Alemi, I.~Fischer, J.~V. Dillon, and K.~Murphy, ``{Deep Variational
  Information Bottleneck},'' \emph{arXiv.org}, 2016.

\bibitem{Pathak:2016vz}
D.~Pathak, P.~Krahenbuhl, J.~Donahue, T.~Darrell, and A.~A. Efros, ``{Context
  Encoders: Feature Learning by Inpainting},'' \emph{arXiv.org}, 2016.

\bibitem{Isola:2016tp}
P.~Isola, J.-Y. Zhu, T.~Zhou, and A.~A. Efros, ``{Image-to-Image Translation
  with Conditional Adversarial Networks},'' \emph{arXiv.org}, 2016.

\bibitem{Li:2019wg}
Y.~Li and S.~Ji, ``{L0-ARM: Network Sparsification via Stochastic Binary
  Optimization},'' in \emph{ECML}, 2019.

\bibitem{Yin:2019tm}
M.~Yin and M.~Zhou, ``{ARM - Augment-REINFORCE-Merge Gradient for Stochastic
  Binary Networks.}'' in \emph{ICLR}, 2019.

\bibitem{Kingma:2014tz}
D.~P. Kingma and M.~Welling, ``{Auto-Encoding Variational Bayes},'' in
  \emph{ICLR}, 2014.

\bibitem{Higgins:2017vm}
I.~Higgins, L.~Matthey, A.~Pal, C.~Burgess, X.~Glorot, M.~Botvinick,
  S.~Mohamed, and A.~Lerchner, ``{beta-VAE - Learning Basic Visual Concepts
  with a Constrained Variational Framework.}'' \emph{ICLR}, 2017.

\bibitem{Rezende:wu}
D.~J. Rezende and F.~Viola, ``{Generalized ELBO with Constrained Optimization,
  GECO},'' \emph{bayesiandeeplearning.org}, 2018.

\bibitem{Rezende:2018wh}
------, ``{Taming VAEs},'' \emph{arXiv.org}, 2018.

\bibitem{Molchanov:2017wh}
D.~Molchanov, A.~Ashukha, and D.~Vetrov, ``{Variational Dropout Sparsifies Deep
  Neural Networks},'' in \emph{ICML}, 2017.

\bibitem{Li:2017th}
H.~Li, A.~Kadav, I.~Durdanovic, H.~Samet, and H.~P. Graf, ``{Pruning Filters
  for Efficient ConvNets},'' in \emph{ICLR}, 2017.

\bibitem{Dai:2018vp}
B.~Dai, C.~Zhu, and D.~Wipf, ``{Compressing Neural Networks using the
  Variational Information Bottleneck},'' in \emph{ICML}, 2018.

\bibitem{Louizos:2018tl}
C.~Louizos, M.~Welling, and D.~P. Kingma, ``{Learning Sparse Neural Networks
  through L0 Regularization},'' in \emph{ICLR}, 2018.

\bibitem{Williams:1992fi}
R.~J. Williams, ``{Simple Statistical Gradient-Following Algorithms for
  Connectionist Reinforcement Learning},'' \emph{Machine Learning}, 1992.

\bibitem{Bengio:2013vv}
Y.~Bengio, N.~L{\'e}onard, and A.~Courville, ``{Estimating or Propagating
  Gradients Through Stochastic Neurons for Conditional Computation},''
  \emph{arXiv.org}, 2013.

\bibitem{Bird:2018uz}
T.~Bird, J.~Kunze, and D.~Barber, ``{Stochastic Variational Optimization},''
  \emph{arXiv.org}, 2018.

\bibitem{Li:2017tha}
H.~Li, A.~Kadav, I.~Durdanovic, H.~Samet, and H.~P. Graf, ``{Pruning Filters
  for Efficient ConvNets},'' in \emph{ICLR}, 2017.

\bibitem{Liu:2019vv}
Z.~Liu, M.~Sun, T.~Zhou, G.~Huang, and T.~Darrell, ``{Rethinking the Value of
  Network Pruning},'' in \emph{ICLR}, 2019.

\bibitem{dsprites17}
L.~Matthey, I.~Higgins, D.~Hassabis, and A.~Lerchner, ``dsprites:
  Disentanglement testing sprites dataset,''
  https://github.com/deepmind/dsprites-dataset, 2017.

\bibitem{3dshapes18}
C.~Burgess and H.~Kim, ``3d shapes dataset,''
  https://github.com/deepmind/3dshapes-dataset, 2018.

\bibitem{Reed:2015uw}
S.~E. Reed, Y.~Zhang, Y.~Zhang, Y.~Zhang, and H.~Lee, ``{Deep Visual
  Analogy-Making},'' in \emph{NIPS}, 2015.

\end{thebibliography}

\newpage



%\vfill


\end{document}